\documentclass[runningheads]{llncs}
\usepackage[T1]{fontenc}
\usepackage{tabularx}
\usepackage{hyperref}

\usepackage{listings}
\usepackage{tcolorbox}
\tcbuselibrary{listings}

\newtcblisting{appendixlisting}{
    colback=white,
    colframe=black,
    boxrule=0.5pt,
    arc=2pt,
    boxsep=0pt,
    left=5pt,
    right=5pt,
    top=5pt,
    bottom=5pt,
    listing only,
    listing options={
        basicstyle=\ttfamily\scriptsize,
        breaklines=true,
        columns=fullflexible
    }
}

\newtcblisting{summarylisting}{
    colback=light_gray,
    colframe=black,
    boxrule=0.5pt,
    arc=2pt,
    boxsep=0pt,
    left=5pt,
    right=5pt,
    top=5pt,
    bottom=5pt,
    listing only,
    listing options={
        basicstyle=\ttfamily\scriptsize,
        breaklines=true,
        columns=fullflexible
    },
    caption=Training Sample with Summary,
    label=summary_lst
}

\usepackage{graphicx}
\begin{document}

\title{Leveraging Large Language Models in Code Question Answering: Baselines and Issues}
\titlerunning{Leveraging LLMs in Code Q\&A: Baselines and Issues}
\author{
% Georgy Andryushchenko\orcidID{0009-0004-4088-2543} \and 
Georgy Andryushchenko \and 
% Vladimir Ivanov\orcidID{0000-0003-3289-8188} \and 
Vladimir Ivanov \and 
% Vladimir Makharev\orcidID{0000-0003-0319-4042} \and \\
Vladimir Makharev \and \\
Elizaveta Tukhtina \and %\orcidID{2222--3333-4444-5555} \and \\
% Aidar Valeev\orcidID{0000-0002-9511-1488}
Aidar Valeev
}
\authorrunning{G. Andryushchenko et al.}
\institute{
Innopolis University, Innopolis, 420500, Russia\\
\email{georgyandryuschenko@gmail.com, v.makharev@innopolis.university, \{v.ivanov, ai.valeev, e.tukhtina\}@innopolis.ru}
}

\maketitle              % typeset the header of the contribution
\begin{abstract}
Question answering over source code provides software engineers and project managers with helpful information about the implemented features of a software product. This paper presents a work devoted to using large language models for question answering over source code in Python. The proposed method for implementing a source code question answering system involves fine-tuning a large language model on a unified dataset of questions and answers for Python code. To achieve the highest quality answers, we tested various models trained on datasets preprocessed in different ways: a dataset without grammar correction, a dataset with grammar correction, and a dataset augmented with the generated summaries. The model answers were also analyzed for errors manually. We report BLEU-4, BERTScore F1, BLEURT, and Exact Match metric values, along with the conclusions from the manual error analysis. The obtained experimental results highlight the current problems of the research area, such as poor quality of the public genuine question-answering datasets. In addition, the findings include the positive effect of the grammar correction of the training data on the testing metric values. The addressed findings and issues could be important for other researchers who attempt to improve the quality of source code question answering solutions. The training and evaluation code is publicly available\footnote{\url{https://github.com/IU-AES-AI4Code/CodeQuestionAnswering}}.

\end{abstract}
\section{Introduction}

Generative AI plays a significant role in software engineering, and Code Question Answering is an emerging application within this domain. Question Answering over source code involves generating textual answers to code-related questions. These questions can be formulated by software engineers, project managers, or even by the other generative model. Integrating Generative AI in the form of answering relevant questions about code can bring tangible value to software teams.

\begin{figure}[!tbh]
    \centering
    \includegraphics[width=\linewidth]{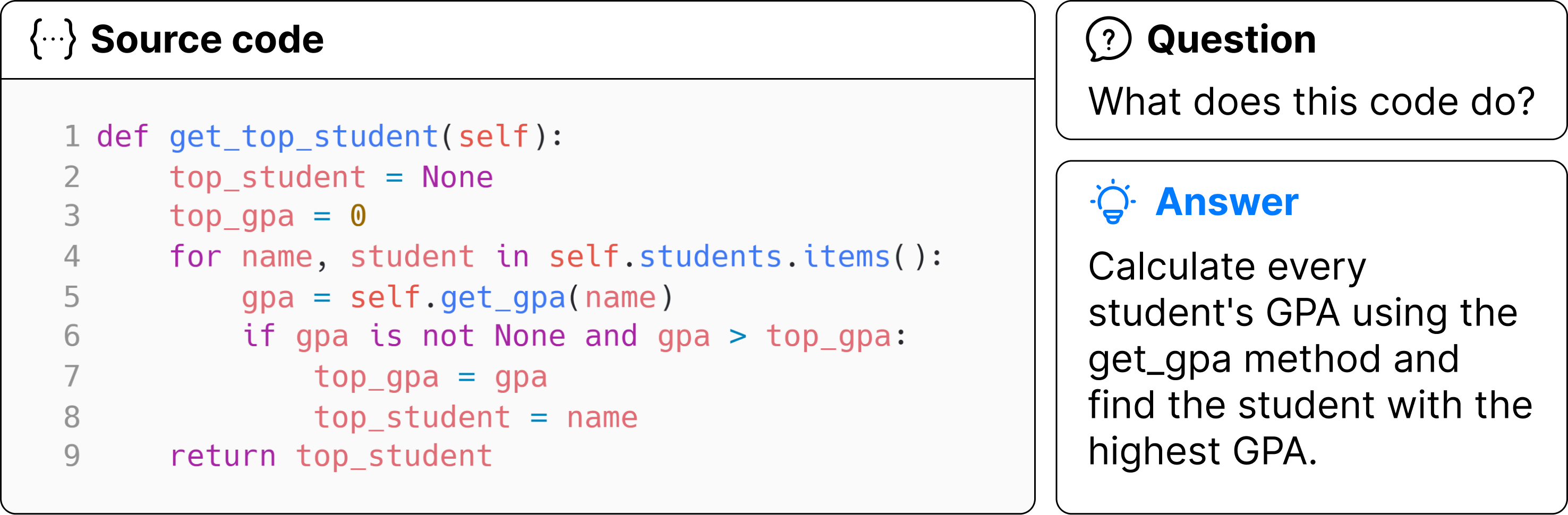}
    \caption{Example of Code Q\&A task: a question and its corresponding answer on source code.}
    \label{fig:codeqa-example}
\end{figure}

Code Question Answering (Code Q\&A) refers to a task in natural language processing and software engineering (see Figure \ref{fig:codeqa-example}). Unlike traditional code comments or documentation, Code Q\&A focuses on interpreting and responding to specific queries about the codebase \cite{CodeQA Dataset}.

In the Code Q\&A scenario, a model is trained to understand natural language questions on a given code piece and to generate appropriate responses. This approach enhances code comprehension by providing developers and stakeholders with an interactive and context-aware way for engaging with the code.

By leveraging Code Q\&A, software teams can efficiently obtain answers to the queries about their codebase. This approach is versatile, accommodating questions from both developers and project managers. AI-powered Code Q\&A streamlines the process and enhances communication within the software development lifecycle whether the software teams seek clarification on code implementation or understanding the software project progress,

Despite current advancements in Large Language Models (LLMs), addressing code question answering remains tough. Indeed, although LLMs show promise as intelligent assistants, evaluating LLM performance is challenging. Existing datasets and benchmarks are not enough to assess Code Q\&A systems since they focus on specific domains and concepts. Human evaluation of such systems would be the best approach, but it requires a lot of resources, such as human annotators. 

Another major challenge is the lack of  high-quality data for code question answering. While certain open-source LLMs exhibit competence as Code Q\&A assistants in zero-shot mode, fine-tuning these models can substantially enhance their performance. Notably, the drawback of many existing datasets for this task lies in the absence of genuine user data since most of these datasets are artificially generated and incorporate no real user inputs. An additional limitation of these datasets lies in their focus on function or method level question answering. For a comprehensive understanding of code, one has to consider the context of entire code files or even repositories.

In this research study, we conducted a series of experiments aimed at assessing the capabilities of LLMs in code question answering, with a specific emphasis on open-source models for code, such as StarCoder \cite{starcoder} and DeepSeek-Coder \cite{deepseekcoder}. Therefore, as a research question, we are striving to determine which data processing and modelling approaches allow to improve the evaluation metrics of the code question-answering system based on LLM. In order to reserach this question we built the baseline and tried to improve its evaluation metrics.

\section{Related Work}

Advancements in natural language processing have significantly impacted code representation, analysis, and generation. This progess focuses on transformer-based models, which treat source code as token streams enriched with structural information. At inference, these models generate outputs autoregressively, producing text, code, or both. Notably, methods such as Code2Text have emerged  in automatic code documentation. Such methods focused on code summarization, comment generation, and question answering over source code. This focus leverages large-scale pre-trained models to understand and interact with code, supporting applications such as legacy code comprehension and code snippet evaluation. Novel approaches in this domain use agent-driven LLMs for repository-level documentation generation \cite{RepoAgent}.

Several state-of-the-art datasets and models have contributed to this field. For example, the CodeQA dataset \cite{CodeQA Dataset} includes Java and Python question-answer pairs derived from code comments. This inclusion facilitates code comprehension through models such as CodeBERT \cite{CodeBERT} and Transformer-based architectures. Some other examples include the CS1QA \cite{CS1QA} dataset, originating from an introductory programming course, and the CodeQueries dataset \cite{CodeQueries}, with its diverse set of queries and spans. These datasets provide valuable resources for training models on code-related questions. In addition, CodeBERT \cite{CodeBERT}, RoBERTa \cite{RoBERTa}, and GraphCodeBERT \cite{GraphCodeBERT} have demonstrated their effectiveness in understanding and generating code-related responses by integrating semantic and syntactic features into their training.

Additionally, techniques for code summarization and representation have seen significant improvements. Indeed, several approaches have incorporated various pre-training objectives thus enchancing the ability to generate accurate and meaningful summaries of code. Examples of such approaches include Code2Seq \cite{Code2Seq} and DeepCom \cite{DeepCom} that use code graphs for embedding representations, as well as multi-task models such as CodeT5 \cite{CodeT5} and SPT-Code \cite{SPTCode}. Meanwhile, methods for parameter-efficient fine-tuning, including prompt tuning and adapters, offer efficient ways to adapt large pre-trained models to specific Code Q\&A tasks without requiring extensive computational resources. Together, these developments establish a robust foundation for exploring and improving Code Q\&A with LLMs in practice.

\section{Datasets}

% In this section we will describe training and testing datasets we used for experiments and how we composed them.

\subsection{Training Datasets}

\subsubsection{Unified Dataset}

% In order to train a question-answering model over code we composed a unified dataset\footnote{\url{https://huggingface.co/datasets/datapaf/UltimateQAFiltered2}} consisting of two public datasets: CodeQA \cite{CodeQA Dataset} and CS1QA \cite{CS1QA}. The original CodeQA dataset consists of functions, questions and answers generated from a code-comments corpus by syntax parsing. In the CodeQA the answers are short, either a single word or a phrase. The CS1QA consists of code gathered from an online introductory programming course. The code goes along with a chat log, a question, an answer, and type of the question. However, the initial quality of the two datasets is not perfect since the datasets contains grammatical errors and noisy examples, e.g., answers suggesting to search for an answer on the internet. While exploring the data, we discovered examples where the code, questions, or answers were deviating from the majority in both content and structure. We removed the examples with undesirable formatting. Some of the answers in the unified dataset have quite long answers, frequently separated into multiple lines. This has led to ambiguity for the model in determining when to provide a longer answer. Consequently, we removed the answers that exceed four lines. Furthermore, certain answers within the unified dataset include irrelevant code and information. Such examples may confuse the model too and negatively affect the effectiveness of the model training. These examples were found by the search by specific keywords and excluded from the dataset. After all these manipulations we got a unified dataset.

To train a question-answering model on code, we compiled a unified dataset\footnote{\url{https://huggingface.co/datasets/datapaf/CodeQuestionAnswering}} from the public datasets: CodeQA \cite{CodeQA Dataset}, CS1QA \cite{CS1QA} and a subset of PCSD \cite{PCSD}. Given Python's clear syntax and its widespread use in academia and industry, we focused exclusively on Python code from both datasets to enhance the relevance of our research. Our dataset choices were limited by the availability of suitable public data.

CodeQA comprises functions, questions, and short answers generated from a code-comments corpus using syntax parsing, with subsets in both Java and Python. The dataset contains "What", "Where", "When", "How", "Why" questions ("Wh-questions") and Yes/No questions. The detailed information about CodeQA dataset may be found in the original paper.

CS1QA contains Python code sourced from an online introductory programming course, including chat logs, questions, answers, and question types. The authors of the dataset split the questions in the dataset into the following types: "variable", "code understanding", "reasoning", "code explain", "error", "algorithm", "usage", "task", "logical". The original paper provides the other details of the dataset.

PCSD comprises the samples of Python code with their summaries. Since this dataset contains only the code summaries, all samples have the same question: "What does this code do?".

The initial datasets contained grammatical errors and noisy examples, such as answers suggesting internet searches. During data exploration, we found cases where the code, questions, or answers were poorly aligned in content or structure, and removed examples with undesirable formatting. Some answers were excessively long, spanning multiple lines, which could lead to ambiguity in model responses, so we excluded those exceeding four lines. Certain types of questions, particularly, "error", "algorithm", "usage", "task", and "logical" were excluded from the unified dataset to avoid potential factual inaccuracies. We also eliminated answers containing irrelevant code or information, identified by specific keywords (see Appendix \ref{apx:a4_keywords} for details), to prevent confusion and improve training effectiveness. 
% As the result of the filtering, 12\% of the initial dataset The distribution of filtered samples is summarized in Table \ref{tab:unified}.
After filtering 12\% of the initial data, we derived a unified dataset, presented in Table \ref{tab:unified}.
% summarizes the statistics of filtered samples, with 12\% of the initial data being filtered out.

% In the unified dataset, we joined Python part of CodeQA \cite{CodeQA Dataset}, CS1QA \cite{CS1QA} and a subset from PCSD \cite{PCSD} with all the samples having a question "What does this code do?". The distributions of the samples before filtering is presented in Table \ref{tab:unified}.

% \begin{table}[ht]
%     \centering
%     \caption{Unified Dataset Samples Distribution before Filtering}
%     \setlength{\tabcolsep}{12pt} % Adjust column separation
%     \renewcommand{\arraystretch}{1.2} % Adjust row separation
%     \begin{tabular}{|l|c|c|c|}
%         \hline
%         \textbf{Source} & \textbf{Train} & \textbf{Validation} & \textbf{Test} \\
%         \hline
%         CodeQA & 56085 & 7000 & 7000 \\
%         \hline
%         CS1QA & 5543 & 1847 & 1847 \\
%         \hline
%         PCSD & 10000 & 1000 & 1000 \\
%         \hline
%     \end{tabular}
%     \label{tab:unified}
% \end{table}
% total 91322

\begin{table}[ht]
    \centering
    \caption{Unified Dataset split for train, validation, and test.}
    \setlength{\tabcolsep}{12pt}
    \begin{tabular}{|l|r|r|r|}
        \hline
        \textbf{Source} & \textbf{Train} & \textbf{Validation} & \textbf{Test} \\
        \hline
        CodeQA & 56,081 & 6,998 & 6,997 \\
        \hline
        CS1QA & 2,035 & 674 & 683 \\
        \hline
        PCSD & 5,660 & 556 & 548 \\
        \hline
        % \bottomrule
        % \hline
        \textit{Total} & 63,776 & 8,228 & 8,228\\
        \hline
    \end{tabular}
    \label{tab:unified}
\end{table}
% total 80232 (12% filtered)

\subsubsection{Unified Dataset with Grammatical Corrections}

Given that the dataset contains grammatical errors, these errors should be naturally corrected to see whether the results improve. Therefore, we applied Language Tool\footnote{\url{https://languagetool.org/}} to identify the grammatical errors in questions. However, we did not correct the answers because it would have involved correcting the reference answers, and thus, making the comparison less fair. Finally, we classified the grammatical errors using the Python wrapper for Language Tool, manually labeled the error types that occur more than twice whether corrected or not, and automatically corrected the selected types if the tool suggested a correction. The dataset is publicly accessible at Hugging Face\footnote{\url{https://huggingface.co/datasets/datapaf/UltimateQAGrammarCorrected}}.

\subsubsection{Unified Dataset with Generated Summaries}

We hypothesized that including short textual summaries of the code may enhance answer. To develop an effective summarization model, we focused on utilizing the CodeT5+ \cite{CodeT5} model with 220 million parameters. The weights of this model are publicly available\footnote{\url{https://huggingface.co/datapaf/CodeT5Summarization}}. For training and evaluation, the Docstring \cite{docstring} dataset was used. This dataset consisted of a collection of summaries for Python functions. After the summarization model was developed, we generated the summaries for the code samples of the unified question answering dataset. These summaries were added to the samples of the unified dataset, providing additional contextual information about the code\footnote{\url{https://huggingface.co/datasets/datapaf/UltimateQASummaries}}.

\subsection{Testing Datasets}

\subsubsection{Testing Subset of Unified Dataset}

We used the unified dataset as one of our testing datasets, and extracted a subset that was not included in the training phase, i.e. the testing subset. This testing dataset comprised 8,228 samples after filtering. Refer to Appendix \ref{apx:a1_unified_testing} for a sample example.

% As one of our testing datasets, we utilized the original unified dataset and extracted a subset that was not included in the training phase.  This testing subset comprises 8,228 samples after filtering. Refer to Appendix \ref{apx:a1_unified_testing} for a sample example.

\subsubsection{Testing Dataset Based on ClassEval Dataset (ClassEvalQA)}

% As another testing dataset we created a dataset consisting of code that was excluded from both the pre-training and fine-tuning phases of the models. For the ClassEval dataset, the authors manually devised a collection of 100 Python coding tasks at the class level and provided human-written solutions. The use of unique human-written code in the ClassEval dataset assures that an LLM has not seen this code in its pre-training phase. Utilizing state-of-the-art LLMs for data generation is a widely used approach when human resources are limited. To create a benchmark for evaluating our Code Q\&A models at the class level, we generated question-answer pairs using GPT-3.5\footnote{\url{https://openai.com/index/gpt-3-5-turbo-fine-tuning-and-api-updates/}}. Approximately 20 pairs were generated for each solution from ClassEval. We instructed GPT-3.5 to generate method-based questions, incorporating information from the entire class. The resulting benchmark\footnote{\url{https://huggingface.co/datasets/datapaf/ClassEvalQABenchmark}} contains 2,050 question-answer pairs and can be used for both function and class-level code question-answering. Refer to Appendix \ref{apx:a2_classeval} for a sample example.

% Vladimir ------------------------
As an additional testing dataset, we created a dataset consisting of code excluded from both the pre-training and fine-tuning phases of the models that we experimented with. The ClassEval dataset \cite{classeval} was manually curated 100 Python coding tasks at the class level, accompanied by human-written solutions. This ensures that the code in ClassEval has not been encountered by the LLM during its pre-training phase. Given the constraints of human resources, leveraging state-of-the-art LLMs like GPT-3.5 for data generation is a common practice. To establish a benchmark for evaluating our Code Q\&A models at the class level, we generated question-answer pairs using GPT-3.5\footnote{\url{https://openai.com/index/gpt-3-5-turbo-fine-tuning-and-api-updates/}}, producing approximately 20 pairs per solution from ClassEval. Specifically, we instructed GPT-3.5 to generate method-based questions that incorporate information from the entire class. The resulting benchmark ClassEvalQA\footnote{\url{https://huggingface.co/datasets/datapaf/ClassEvalQABenchmark}} contained 2,050 question-answer pairs, suitable for both function and class-level code question-answering. For a sample example, see Appendix \ref{apx:a2_classeval}.
% Vladimir ------------------------

\subsubsection{High Quality Subset of Unified Dataset}

% Manual checking of the samples has shown that testing subset of the unified dataset has quality problems. These problems include grammatical mistakes, unclear formulation of the question, irrelevant answer, excessively wordy answer. They may threaten the validity of the experiments. In order to avoid such threats, we considered a dataset which is the subset of the initial testing dataset that was manually composed by selecting 100 examples of higher quality\footnote{\url{https://huggingface.co/datasets/datapaf/UCQAQualitySubsetBenchmark}}. Higher quality of the examples means that they do not have the quality problems described above. For a sample example, see Appendix \ref{apx:a3_hq_subset}.

% Vladimir ------------------------------
Manual inspection of the testing subset from the unified dataset revealed quality issues, including grammatical errors, unclear question formulations, irrelevant answers, and overly verbose answers. Such problems might threaten the validity of our experiments. To mitigate this risk, we curated a subset\footnote{\url{https://huggingface.co/datasets/datapaf/UCQAQualitySubsetBenchmark}} of the initial testing dataset, manually selecting 100 high-quality examples that do not exhibit the aforementioned quality problems. For a sample example, see Appendix \ref{apx:a3_hq_subset}.
% Vladimir ------------------------------

\section{Modeling}
% Georgy

LLMs demonstrate an impressive power of text understanding and generalization \cite{Brown}. The pre-training procedure makes the LLMs understand a vast set of source code snippets. 
% Since the questions may significantly vary in their formulation and essence, we decided to consider open-source LLMs for their capabilities. 
These code snippets can be written in different ways and connected to different domains, such as backend, data science, game development, etc. Thus LLMs can perceive highly varying code.

As the models for experimenting we have chosen StarCoder \cite{starcoder} and DeepSeek-Coder \cite{deepseekcoder}. StarCoder\footnote{\url{https://huggingface.co/bigcode/starcoder}} has a GPT-2 architecture with 15.5 billion parameters. It was pre-trained on a dataset of 1 trillion tokens. DeepSeek-Coder model is a decoder-only Transformer. We selected the 6.7 billion model version\footnote{\url{https://huggingface.co/deepseek-ai/deepseek-coder-6.7b-instruct}}. DeepSeek-Coder was pre-trained made on a dataset of 2 trillion tokens collected by the authors of the model.

We fine-tuned StarCoder and DeepSeek-Coder with the qLoRA \cite{qLoRA} approach with various low-rank adaptation (LoRA) hyperparameters. Usually, the fine-tuning of the model took about 24 hours on a single A100 GPU with 80 GB of video memory. We fine-tuned all the models on the different versions of the dataset: Unified Dataset, Unified Dataset with Grammar Correction, and Unified Dataset with Summaries.

\section{Evaluation}
% Aidar
This section describes the metrics that we used to evaluate our fine-tuned versions of StarCoder and DeepSeek-Coder models on the collected datasets. In addition, this section discusses the model results and provides error analysis.

\subsection{Metrics}
To compare the generated text with the true answer we evaluated the values of the following automated metrics:
% We used an implementation from HuggingFace library evaluate-metric.

\begin{itemize}
    \item \textbf{BLEU} \cite{BLEU} is an algorithm to compare the similarity between generated and reference sentences. The score represents the $n$-gram precision between the machine-generated sentence and the reference translations with a brevity penalty. We chose $n = 4$ since the metric correlates well with human judgements with such $n$. BLEU scores range from 0 to 1, where the perfect match of the candidate with the reference was indicated by 1.
    \item \textbf{BERTScore} \cite{BERTScore} is a semantic similarity evaluation metric for text generation that leverages the embeddings from pre-trained BERT \cite{bert} model. This metric compares machine-generated and reference sentences using cosine similarity, and considers the cases with synonyms. As a backbone model we used \texttt{microsoft/deberta-xlarge-mnli} checkpoint as recommended by the authors of the metric. BERTScore produces three values: Precision, Recall, and F1. The F1 value was considered since it includes both Precision and Recall values. All the values ranged from 0 to 1, where 1 was the prefect result. 
    \item \textbf{BLEURT} \cite{bleurt} is a BERT-based evaluation metric for text generation. The metric uses a BERT-based regression model trained on publicly available collection of ratings to obtain the value. As a backbone model we used \texttt{BLEURT-20} checkpoint as recommended by the authors of the metric. The metric values were mostly between 0 and 1, where 1 indicates the best similarity. However, sometimes the values may be below 0 and over 1, as the authors of the metric state.
    \item \textbf{F1-score ($F_1$)} is the harmonic mean of token-based precision and recall. Precision is the fraction of common tokens in the candidate that appear in the reference text, while recall is the fraction of common tokens in the reference text that appear in the candidate. The F1-score ranged from 0 to 1 where 1 means the highest correspondence of the candidate to the reference. 
    \item \textbf{Exact Match (EM)} calculates the percentage of generated answers that exactly match the reference answers. The higher the percentage, the more similar the candidate is to the reference.
\end{itemize}

\begin{figure}[!tbh]
    \centering
    \includegraphics[width=0.6\linewidth]{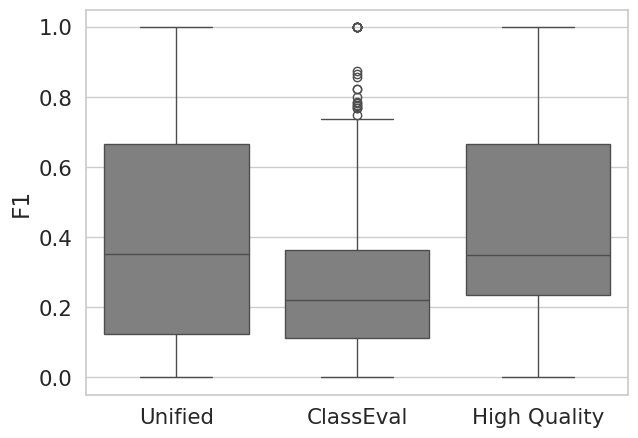}
    \caption{Distribution of the token-wise F1-score values across three test datasets.}
    \label{fig:f1-distribution}
\end{figure}

We chose BLEU-4, BERTScore F1, BLEURT metrics for our experiments based on their popularity and robustness \cite{nlg_metrics_survey}. In addition to these metrics, we considered F1-score and Exact match scores that are often used in question answering tasks \cite{qa_metrics_survey}. However, when analyzing the results, we excluded F1-score from the further analysis due the high deviation of the values as shown in Figure \ref{fig:f1-distribution}.

\begin{figure}[!tbh]
    \centering
    \includegraphics[width=0.6\linewidth]{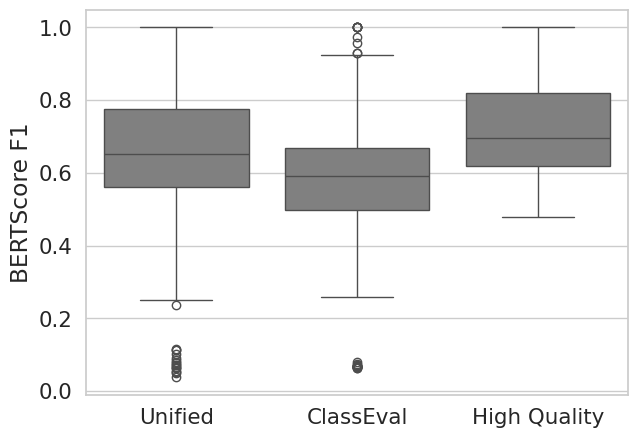}
    \caption{Distribution of the BERTScore F1 values across three test datasets.}
    \label{fig:bertscore-f1-distribution}
\end{figure}

Note that in our experiments we calculated the average values only for BERTScore F1 and BLEURT metrics, while BLEU-4 and Exact Match values were calculated over the entire testing set.

\subsection{Results}

This subsection discusses the results of the experiments regarding the model hyperparameters. The subsection provides detailed results with the metrics values obtained from the experiments.

\subsubsection{Model Hyperparameters}
StarCoder and DeepSeek-Coder were trained with different sequence length and LoRA parameters. Evaluating the training results showed that the optimal sequence length was 1024, the optimal LoRA rank was 32 with LoRA alpha 64, and the dropout was 0.05 for both models. 

\subsubsection{Decoding Strategy}
We also tried different decoding strategies to find out whether these strategies affected the quality of the generated answers. The tried decoding strategies were the following: greedy, sampling, sampling with temperature 0.6, top three sampling, sampling with top probability 0.99. Our experiments showed, that changing the decoding strategy decreased the quality of the answers comparing to the originally proposed approach with the greedy strategy. \\

The final results of our experiments are presented in Table \ref{tab1}, Table \ref{tab2}, and Table \ref{tab3}. Note that the average values of BLEURT and BERTScore F1 metrics are supplied with the values of standard deviation to describe the spread of the values.
Table \ref{tab1} compares two ways to data augmentation for different models: adding generated summaries and grammar correction in questions.

\begin{table}[ht!]
\caption{Evaluation results on the testing subset of the unified dataset (in percentage).}
\label{tab1}
\begin{tabularx}{\textwidth}{|>{\raggedright\arraybackslash}p{81pt}|>{\centering\arraybackslash}p{70pt}|>{\centering\arraybackslash}p{80pt}|>{\centering\arraybackslash}p{40pt}|>{\centering\arraybackslash}X|}
\hline
\textbf{Model} &  \textbf{BLEURT} & \textbf{BERTScore F1} & \textbf{BLEU-4} & \textbf{EM}\\
\hline
StarCoder  & 40.04 $\pm$ 26.21 &	65.89 $\pm$ 16.88 & 7.42 & 9.89 \\
+Summaries & 36.85 $\pm$ 25.46 &	64.20 $\pm$ 16.31 & 5.63 & 8.34 \\
+Grammar &  \textbf{42.87} $\pm$ \textbf{25.96} & \textbf{67.45} $\pm$ \textbf{16.57} & \textbf{8.89} & \textbf{10.83} \\
\hline
DeepSeek-Coder  & 38.42 $\pm$ 26.38 &	65.21 $\pm$ 17.12 & 4.95 &	9.72 \\
+Summaries & 38.63 $\pm$ 25.27 &	65.33 $\pm$ 16.08 & 4.80 &	8.69  \\
+Grammar & 38.29 $\pm$ 24.56 &	65.14 $\pm$ 15.35 & 4.90 &	7.74 \\
\hline
\end{tabularx}
\end{table}

Quite many cases of incorrect predictions were caused by problems with the quality of the testing dataset. These problems include grammatical mistakes, unclear formulation of the question, irrelevant answer, excessively wordy answer. They may threaten the validity of the experiments. To avoid such a threat, the baseline and the proposed models were tested on the other two testing datasets: ClassEvalQA and High Quality testing datasets.
% One of the datasets is the subset of the initial testing dataset that was manually composed by selecting examples of higher quality (the high quality testing subset from the Table \ref{tab3}). Another dataset was collected using the code from ClassEval dataset.
% .: ClassEvalQA and High Quality testing datasets.

\begin{table}[ht!]
\caption{Evaluation results on the ClassEvalQA testing dataset (in percentage). In this table and in Table \ref{tab3}, `+S' denotes the model trained on the dataset augmented with summaries while `+G' denotes the model trained on the dataset with grammar correction.}
\label{tab2}
\begin{tabularx}{\linewidth}{|>{\raggedright\arraybackslash}p{81pt}|>{\centering\arraybackslash}p{70pt}|>{\centering\arraybackslash}p{80pt}|>{\centering\arraybackslash}p{40pt}|>{\centering\arraybackslash}X|}
\hline
\textbf{Model} &  \textbf{BLEURT} & \textbf{BERTScore F1} & \textbf{BLEU-4} & \textbf{EM}\\
\hline
StarCoder+S  & 26.88 $\pm$ 18.31 &	51.65 $\pm$ 15.15 & 1.18 & 0.34 \\
StarCoder+G & \textbf{35.04} $\pm$ \textbf{17.57} & \textbf{57.95} $\pm$ \textbf{12.75} & \textbf{4.24} & 0.39 \\
\hline
DeepSeek-Coder+S &  31.84 $\pm$ 19.22 &	54.55 $\pm$ 15.56 & 2.44 & 0.48 \\
DeepSeek-Coder+G  & 34.86 $\pm$ 18.32 &	57.11 $\pm$ 14.39 & 3.37 &	\textbf{0.53} \\
\hline
\end{tabularx}
\end{table}

\begin{table}[ht!]
\caption{Evaluation results on the high quality testing subset (in percentage).}
\label{tab3}
\begin{tabularx}{\linewidth}{|>{\raggedright\arraybackslash}p{81pt}|>{\centering\arraybackslash}p{70pt}|>{\centering\arraybackslash}p{80pt}|>{\centering\arraybackslash}p{40pt}|>{\centering\arraybackslash}X|}
\hline
\textbf{Model} &  \textbf{BLEURT} & \textbf{BERTScore F1} & \textbf{BLEU-4} & \textbf{EM}\\
\hline
StarCoder+S & 41.92 $\pm$ 22.83 &	66.75 $\pm$ 13.54 & 4.65 & 7.00  \\
StarCoder+G & \textbf{49.12} $\pm$ \textbf{22.26} & \textbf{71.63} $\pm$ \textbf{13.54}  & \textbf{9.73} & \textbf{9.00}  \\
\hline
DeepSeek-Coder+S & 45.46 $\pm$ 21.77 & 68.26 $\pm$ 12.85 & 6.27 & 6.00 \\
DeepSeek-Coder+G & 47.57 $\pm$ 19.53 & 69.49 $\pm$ 11.93 & 5.86 & 5.00 \\
\hline
\end{tabularx}
\end{table}

% \footnotetext[1]{Here, `+S' denotes the model trained on the dataset augmented with summaries while `+G' denotes the model trained on the dataset with grammar correction.}

\section{Discussion of Results and Open Issues}

\subsection{Metrics Analysis}

As Table \ref{tab1} shows fine-tuning StarCoder on a dataset with corrected grammar improved all the metrics, while adding summaries to the model decreased the metrics. The Table \ref{tab2} and Table \ref{tab3} values indicate that grammar correction outperforms summaries generation.

The conclusions are different for DeepSeek-Coder. The metrics in Table \ref{tab1} demonstrate that adding summaries and fixing grammar do not improve the metrics. This absence of improvement might result from the difference in model size. However, Table \ref{tab2} and Table \ref{tab3} show that fixing grammar provide higher metrics than adding summaries.

To analyze the metrics distribution, we built box plots based on the testing metrics values of StarCoder model trained on text with grammar correction. As Figure \ref{fig:f1-distribution} shows, the standard deviation values for F1-score were relatively high. These high values might indicate that the metric happened to be not precise or consistent. Meanwhile, BERTScore F1 metric demonstrated less variance as seen in Figure \ref{fig:bertscore-f1-distribution}. Lesser standard deviation of the metric values might indicate that BERTScore F1 is a more significant metric that could be used for making conclusions.
% In fact, there is a significant amount of examples with low and high scores which may also be an indicator under-fitted model.

\subsection{Error Analysis and Further Improvement}

To determine the reasons of model mistakes, we manually reviewed the model answers on the high quality testing dataset. As a result of this review, we determined the following possible reasons:

\begin{itemize}
    \item \textbf{Unclear question}. The question was formulated vaguely and ambiguously, so the resulting answer may not be determined.
    \item \textbf{Code lacks necessary information}. Answering the question required the information that the code did not contain.
    \item \textbf{Irrelevant true answer}. In fact, the true answer did not correspond to the question.
    \item \textbf{External usage question}. The question corresponded to the possible usages of the provided code. For example, the question might ask in which cases the code may be useful.
    \item \textbf{Redundant information in the true answer}. The true answer contains the redundant text unrelated to the question.
\end{itemize}

Our findings revealed several common issues that affected the quality of the answers. The most recurring problems were related to insufficient context in the prompt and unclear question formulation. Such issues may be addressed through prompt improvement.

The primary concern revolves around limited context. In most cases, the code of the function or method was insufficient to provide a comprehensive answers to many relevant questions that one could ask. Hence, due to the lack of context the model failed to provide the correct answer in some cases. The solution for this failure would be focusing on repository-level question answering, which is the main direction for our future work. 

\subsection{Limitations of Our Code Q\&A Solution}

As the final model for the solution, we preferred StarCoder model over DeepSeek-Coder. This preference was due to the metric values in Table \ref{tab1}, Table \ref{tab2}, and Table \ref{tab3}. Generally, StarCoder model happened to have higher values.

The proposed solution currently supports only Python due to lack of data in other programming languages. However, the underlying StarCoder model is a multilingual model, so zero-shot capabilities are possible. However, evaluating such capabilities is left for future work.

Our model inherited the context length of StarCoder model that was equal to 1024 tokens. This length corresponds to approximately 700-800 words for code, question and answer altogether. If we assume the average code line is 15 words long, this assumptions leads to a maximum of 50 lines of code. A well-regarded programming style suggests that functions should be up to 20 lines long, so this context should be enough for function-level question answering. However, classes and files were longer than 50 lines in average, so class- or project-level question answering are left for future work.

Another open issue was the evaluation of the question answering. This issue stemmed from the lack of human-labeled datasets and the weaknesses of n-gram based metrics. Metrics like BLEU, which rely on n-grams, only reward the presence of correct words while disregarding synonyms and alternative expressions of the same meaning. Consequently, these metrics lose their correlation with human judgment beyond a certain threshold.

The final issue is the data quality. As discussed above, one of our datasets was synthesized from the code-summaries dataset, and another one was the history of a chat between students and teaching assistants containing grammatical errors and irrelevant answers. In fact, we obtained the most promising results from the manually collected and curated datasets such as ClassEvalQA and High Quality testing datasets. Therefore, for future studies we need to collect a new human-labeled dataset, as well as feedback for our solution in working scenarios.

ChatGPT\footnote{\url{https://openai.com/chatgpt/}} could be a better baseline for code question-answering. However, our research study was funded with a purpose of developing a solution that could be run locally. This requirement is critical for the vast majority of companies for security and code safety reasons.

Current solution is also subject to general language models issues, such as hallucinations, speaking on sensitive topics, variable-name dependence. The usual working scenarios are not affected by these issues. However, we could easily construct a question that invokes undesirable behavior. These issues are also left for future work.

\section{Conclusion}
Question answering over source code is a tough but highly relevant problem. Indeed, a solution could enhance developers' code comprehension, streamline onboarding, and facilitate working with legacy code. In our approach, we tackled this problem using a classical method: collecting training data, cleaning them up, and fine-tuning pre-trained encoder-decoder and decoder-only models. Preliminary experiments indicate that enhancing the model-generated answers is feasible through several avenues. Simultaneously, experiments reveal that utilizing StarCoder and DeepSeek-Coder requires modest initial training set, but the quality of the data remains crucial. Therefore, future steps aim at refinement of the data collection and development of annotation tools. 
% Generating data with LLMs can address both factors effectively. 

\section*{Acknowledgments}
% The authors would like to thank The Advanced Engineering School of Innopolis University for support of this work.

% The authors extend their sincere gratitude to the Schools of Advanced Engineering Studies at Innopolis University for their invaluable provision of resources and facilities, which were essential for the successful execution of this work.

We thank the Advanced Engineering School of Innopolis University for their essential resources and facilities, which were crucial to the project's success. We also appreciate constructive feedback from anonymous reviewers which greatly improved the paper and valuable suggestions on enhancing the text provided by Georgy Gelvanovsky.

\appendix

% \newpage

\section{Testing Datasets and Dataset Preprocessing Details}

\subsection{Unified Testing Dataset Sample}\label{apx:a1_unified_testing}
\begin{appendixlisting}
QUESTION:
What does the code use to parse the metadata from the provided url?

CODE:
def get_data(url):
	try:
		request = requests.get(url)
		request.raise_for_status()
	except (requests.exceptions.HTTPError, requests.exceptions.ConnectionError) as e:
		raise ParseError(e)
	items = microdata.get_items(request.text)
	for item in items:
		if (item.itemtype == [microdata.URI('http://schema.org/Recipe')]):
			return item
		raise ParseError('No recipe data found')

ANSWER:
the metadata module
\end{appendixlisting}

\subsection{ClassEval Sample}\label{apx:a2_classeval}

\begin{appendixlisting}
QUESTION:
What is the purpose of the `filter()` method? 

CODE:
def filter(self, request):
        request_uri = request['path']
        method = request['method']

        if self.is_start_with(request_uri):
            return True

        try:
            token = self.get_jwt_user(request)
            user = token['user']
            if user['level'] > 2:
                self.set_current_user_info_and_log(user)
                return True
        except:
            return False 

ANSWER:
The purpose of the `filter()` method is to filter the incoming request based on certain rules and conditions.

\end{appendixlisting}

\subsection{High Quality Testing Dataset Sample}\label{apx:a3_hq_subset}
\begin{appendixlisting}
QUESTION:
What does the code get ? 

CODE:
def policy_key(location):
	return u'{cat}/{name}'.format(cat=location.category, name=location.name)
 

ANSWER:
the key for a location in a policy file
\end{appendixlisting}

\subsection{Keywords Used to Filter Out Confusing Examples from CodeQA and CS1QA}\label{apx:a4_keywords}
\begin{appendixlisting}
':param', '>>>', 'args:', 'returns:', 'cli example', '@param', ':rtype', ':type', '@type', 'http://', 'https://', 'see also:'
\end{appendixlisting}

\end{document}